\documentclass[letterpaper]{article} % DO NOT CHANGE THIS
\usepackage{aaai2027}  % Camera-ready (remove [submission])
\nocopyright
\usepackage[hyphens]{url}  % DO NOT CHANGE THIS
\usepackage{graphicx} % DO NOT CHANGE THIS
\usepackage{natbib}  % DO NOT CHANGE THIS AND DO NOT ADD ANY OPTIONS TO IT
\usepackage{caption} % DO NOT CHANGE THIS AND DO NOT ADD ANY OPTIONS TO IT
\usepackage{algorithm}
\usepackage{algorithmic}

\usepackage{newfloat}
\usepackage{listings}
\DeclareCaptionStyle{ruled}{labelfont=normalfont,labelsep=colon,strut=off} % DO NOT CHANGE THIS
\floatstyle{ruled}
\newfloat{listing}{tb}{lst}{}
\floatname{listing}{Listing}

\usepackage{booktabs}

\usepackage{tabularx}
\usepackage{array}
\usepackage{makecell}
\usepackage{multirow}
\usepackage{amssymb}
\usepackage{amsmath}
\usepackage{subcaption}
\usepackage{pifont}
\usepackage{xcolor}

\newcommand{\best}[1]{\textbf{#1}}
\newcommand{\second}[1]{\underline{#1}}
\newcommand{\cmark}{\ding{51}}
\newcommand{\xmark}{\ding{55}}
\newcommand{\pmark}{\(\triangle\)}

\newcolumntype{L}[1]{>{\raggedright\arraybackslash}p{#1}}
\newcolumntype{C}[1]{>{\centering\arraybackslash}p{#1}}
\newcolumntype{Y}{>{\raggedright\arraybackslash}X}

\title{SafeSceneReason: A Multimodal Reasoning Benchmark Connecting Industrial Hazards with Accident Knowledge}
\author{
    Yuanchi Zhu\textsuperscript{\rm 2,\rm 5}\equalcontrib,
    Kang An\textsuperscript{\rm 1}\equalcontrib\thanks{Project leader},
    Tengyue Wang\textsuperscript{\rm 6},
    Zhongyu Yang\textsuperscript{\rm 3},
    Chenxu Du\textsuperscript{\rm 7},\\
    Xinqi Yang\textsuperscript{\rm 8},
    Hebao Zhu\textsuperscript{\rm 9},
    Bokai Zhao\textsuperscript{\rm 10},
    Tianyu Liang\textsuperscript{\rm 11},
    Ziliang Wang\textsuperscript{\rm 4},\\
    Faqiang Qian\textsuperscript{\rm 4},
    Yunli Yang\textsuperscript{\rm 12},
    Weiyang Shi\textsuperscript{\rm 2}\corresponding,
    Qibing Ren\textsuperscript{\rm 1}\corresponding
}
\affiliations{
    \textsuperscript{\rm 1}Shanghai Jiao Tong University,
    \textsuperscript{\rm 2}Institute of Automation, Chinese Academy of Sciences,
    \textsuperscript{\rm 3}ModelBest,
    \textsuperscript{\rm 4}SenseTime,\\
    \textsuperscript{\rm 5}ShanghaiTech University,
    \textsuperscript{\rm 6}South China University of Technology,
    \textsuperscript{\rm 7}Southwest Jiaotong University,\\
    \textsuperscript{\rm 8}East China Normal University,
    \textsuperscript{\rm 9}Chongqing University,
    \textsuperscript{\rm 10}University of the Chinese Academy of Sciences,\\
    \textsuperscript{\rm 11}Southeast University,
    \textsuperscript{\rm 12}Institute for Advanced Algorithms Research, Shanghai\\
    zhuych2024@shanghaitech.edu.cn,
    shiweiyang2017@ia.ac.cn,
    \{an\_kang, renqibing\}@sjtu.edu.cn
}

\begin{document}

\maketitle

\begin{abstract}
Industrial-safety understanding requires more than detecting workers, equipment, and personal protective equipment. Models must also assess compliance, identify hazardous interactions, explain potential accident mechanisms, and recommend preventive actions. Existing safety datasets primarily focus on visual perception or isolated violation recognition and provide limited supervision for evidence-grounded reasoning. We introduce \textbf{SafeSceneReason}, a multimodal industrial-safety reasoning benchmark and companion training corpus that connects workplace scenes with knowledge from occupational accident investigations. SafeSceneReason combines two complementary data-construction pipelines. The scene-centric pipeline converts annotated workplace images into executable safety scene graphs and generates deterministic answers through program execution over objects, relations, and safety rules. The report-centric pipeline extracts figures and contextual evidence from accident reports and constructs multimodal questions using evidence graphs, explicit information boundaries, multi-step reasoning paths, and iterative verification. The resulting resource contains 110,581 verified scene-centric question--answer pairs and 13,114 refined report-centric question--answer pairs, covering perception, spatial and quantitative reasoning, compliance assessment, evidence synthesis, causal analysis, and mitigation-oriented decision making. Evaluation of representative proprietary and open-source vision--language models reveals substantial performance differences and persistent weaknesses in comparative, technical, and multi-evidence reasoning, demonstrating that strong general visual understanding does not yet guarantee reliable industrial-safety reasoning.
\end{abstract}

% Keep this block between (not within) the abstract and the main body of the paper.
\begin{links}
    \link{Code}{https://github.com/KangKanng/safescenereason}
    % \link{Datasets}{https://aaai.org/example/datasets}
    % \link{Extended version}{https://aaai.org/example/extended-version}
\end{links}

\begin{table*}[t]
\centering
\caption{
Unified comparison of representative scene-centric safety datasets and
report-centric accident resources.
Most existing works focus on either workplace-scene understanding or
accident-report analysis, whereas SafeSceneReason integrates both sources
for multimodal, multi-step, and evidence-grounded safety reasoning.
\cmark, \pmark, and \xmark\ denote full, partial, and no explicit support,
respectively.
}
\label{tab:unified_dataset_comparison}
\scriptsize
\setlength{\tabcolsep}{3.2pt}
\renewcommand{\arraystretch}{1.15}

\resizebox{\textwidth}{!}{
\begin{tabular}{llccccccccccc}
\toprule
\textbf{Work}
& \textbf{Primary Data}
& \textbf{Scene}
& \textbf{Report}
& \textbf{Compliance}
& \textbf{Risk}
& \textbf{QA}
& \textbf{Multi-hop}
& \textbf{Rationale}
& \makecell{\textbf{Cause--}\\\textbf{Effect}}
& \textbf{Mitigation}
& \makecell{\textbf{Evidence}\\\textbf{Grounding}}
& \makecell{\textbf{Train}\\\textbf{+ Eval}} \\
\midrule

\multicolumn{13}{l}{
\textit{\textbf{Scene-centric visual safety datasets}}
} \\

VQA for Safety Compliance
Checking~\cite{ding2022safety}
& Construction images
& \cmark
& \xmark
& \cmark
& \pmark
& \cmark
& \xmark
& \pmark
& \xmark
& \pmark
& \pmark
& \cmark \\

ConstructionSite 10k~\cite{chen2025constructionsite}
& Construction images
& \cmark
& \xmark
& \cmark
& \pmark
& \cmark
& \pmark
& \cmark
& \xmark
& \pmark
& \cmark
& \cmark \\

IndustryEQA~\cite{li2025industryeqa}
& Simulated warehouse episodes
& \cmark
& \xmark
& \pmark
& \cmark
& \cmark
& \cmark
& \cmark
& \pmark
& \pmark
& \pmark
& \xmark \\

iSafetyBench~\cite{abdullah2025isafetybench}
& Industrial videos
& \cmark
& \xmark
& \pmark
& \cmark
& \xmark
& \xmark
& \xmark
& \xmark
& \xmark
& \xmark
& \xmark \\

SteelBench~\cite{yarrabothula2026steelbench}
& Steel-plant CCTV videos
& \cmark
& \xmark
& \cmark
& \cmark
& \xmark
& \pmark
& \pmark
& \pmark
& \pmark
& \pmark
& \xmark \\

InspecSafe-V1~\cite{liu2026inspecsafe}
& Images and sensor signals
& \cmark
& \xmark
& \pmark
& \cmark
& \xmark
& \xmark
& \pmark
& \xmark
& \pmark
& \pmark
& \cmark \\

\midrule

\multicolumn{13}{l}{
\textit{\textbf{Report-centric accident resources}}
} \\

IncidentAI~\cite{inoue2023safer}
& Accident reports
& \xmark
& \cmark
& \xmark
& \pmark
& \xmark
& \xmark
& \xmark
& \cmark
& \xmark
& \xmark
& \xmark \\

Ahmadi et al.~\cite{ahmadi2025automatic}
& Accident reports
& \xmark
& \cmark
& \xmark
& \pmark
& \xmark
& \xmark
& \xmark
& \pmark
& \xmark
& \xmark
& \xmark \\

Patil et al.~\cite{patil2024improving}
& Accident narratives
& \xmark
& \cmark
& \xmark
& \pmark
& \xmark
& \xmark
& \pmark
& \pmark
& \cmark
& \xmark
& \xmark \\

Chen et al.~\cite{chen2024information}
& Accident reports
& \xmark
& \cmark
& \xmark
& \pmark
& \xmark
& \pmark
& \pmark
& \cmark
& \xmark
& \pmark
& \xmark \\

Liu and Luo~\cite{liu2025large}
& Accident reports
& \xmark
& \cmark
& \xmark
& \pmark
& \xmark
& \pmark
& \pmark
& \cmark
& \pmark
& \pmark
& \xmark \\

\midrule

\textbf{SafeSceneReason (Ours)}
& \textbf{Workplace scenes + accident reports}
& \cmark
& \cmark
& \cmark
& \cmark
& \cmark
& \cmark
& \cmark
& \cmark
& \cmark
& \cmark
& \cmark \\

\bottomrule
\end{tabular}
}
\end{table*}

\begin{figure*}[t]
  \centering
  \includegraphics[width=\textwidth]
      {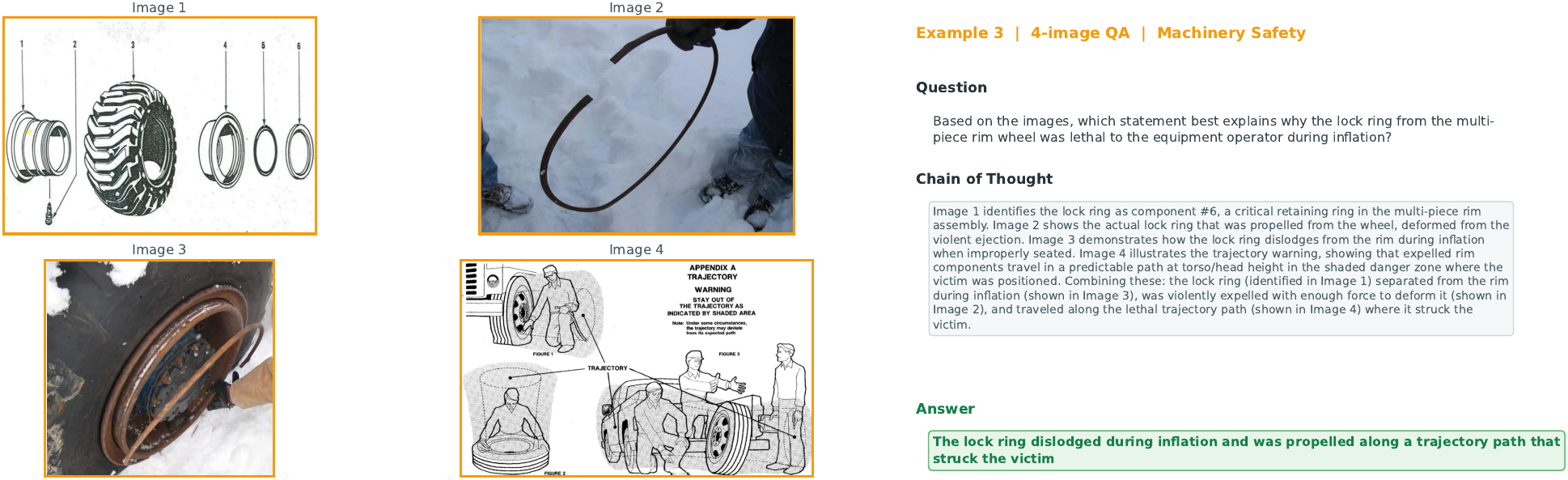}
  \caption{A four-image machinery-safety QA example with its question,
  chain of thought, and final answer.}
  \label{fig:multi-image-machinery-example}
\end{figure*}

\begin{figure*}[t]
  \centering
  \includegraphics[
      width=\textwidth
  ]{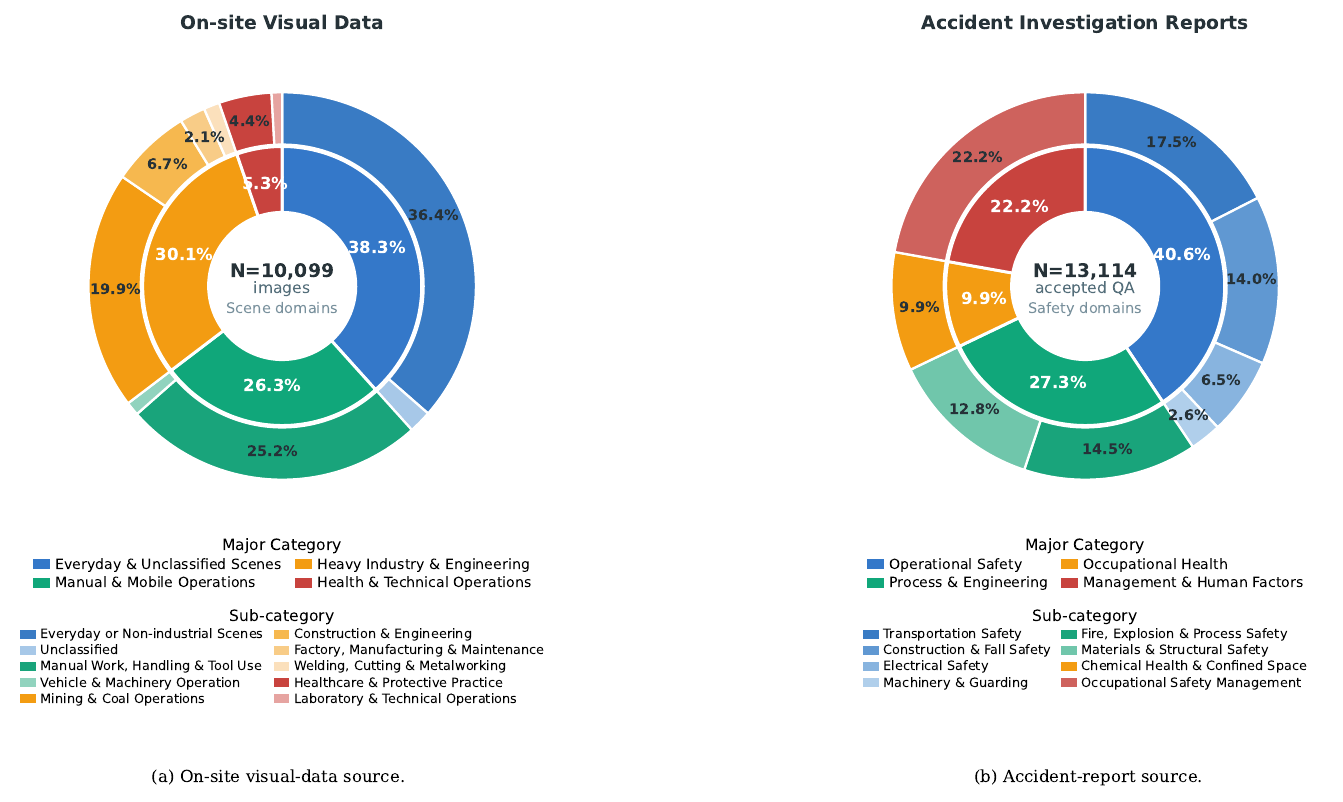}
  \caption{
      Taxonomy distributions of the two industrial-safety data sources.
      (a) On-site visual data.
      (b) Accident investigation reports.
  }
  \label{fig:dual_source_taxonomy}
\end{figure*}

\begin{figure*}[t]
  \centering

  \begin{subfigure}[t]{0.49\textwidth}
      \centering
      \includegraphics[width=\linewidth]
      {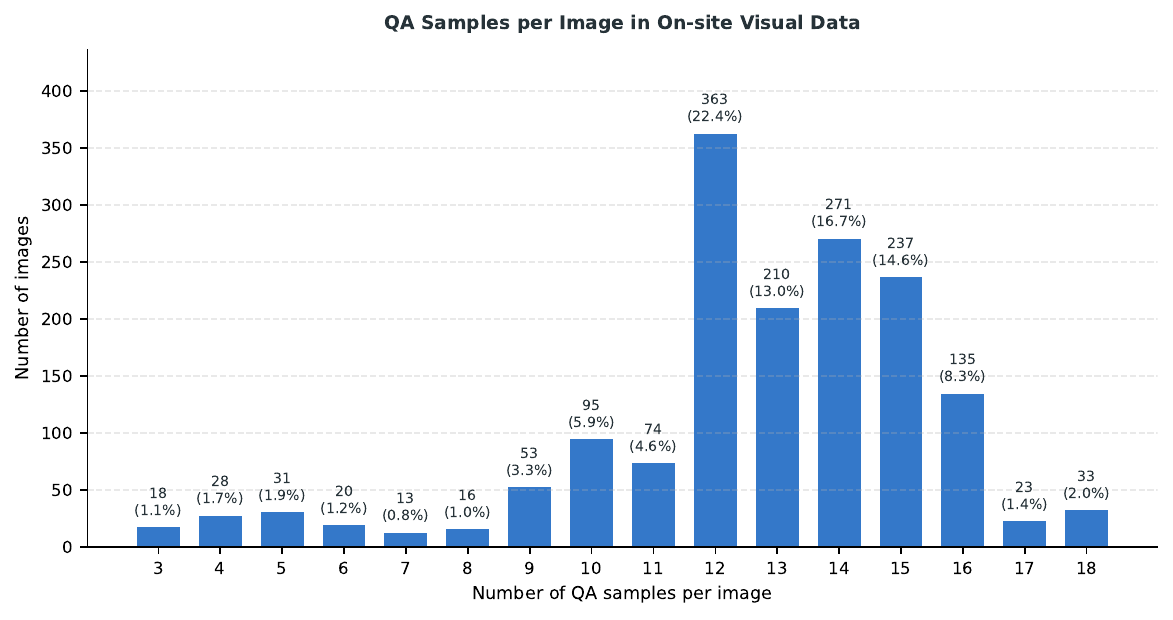}
      \caption{Number of QA samples generated per image in the
      On-site Visual Data.}
      \label{fig:onsite-qa-per-image}
  \end{subfigure}
  \hfill
  \begin{subfigure}[t]{0.49\textwidth}
      \centering
      \includegraphics[width=\linewidth]
      {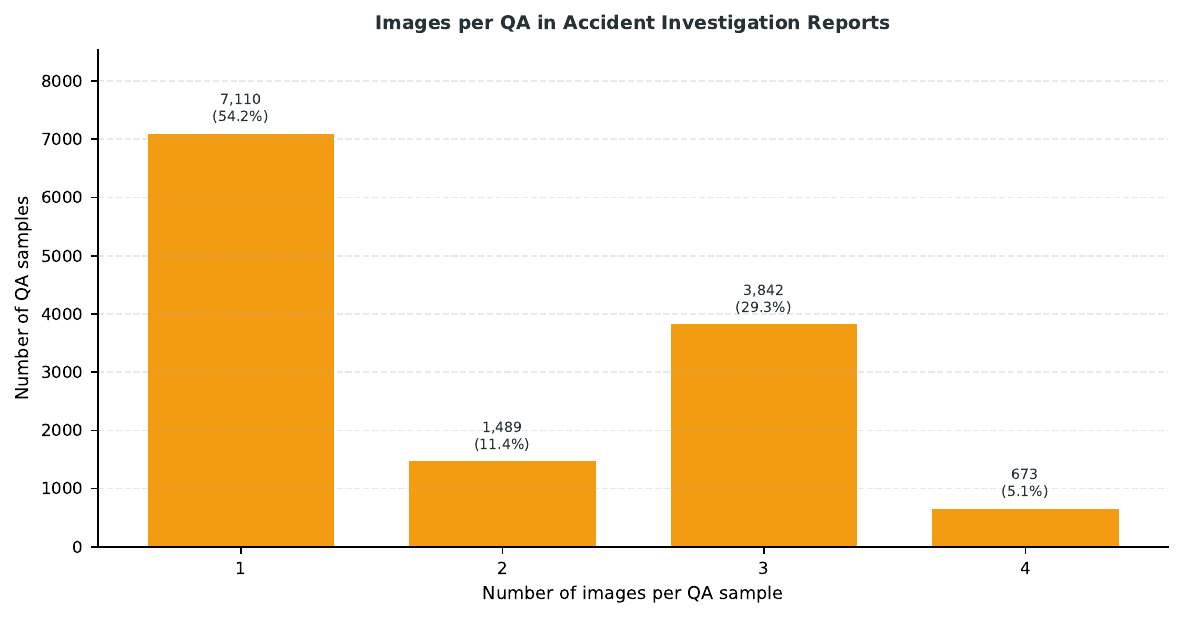}
      \caption{Number of images associated with each QA sample in
      the Accident Investigation Reports.}
      \label{fig:report-images-per-qa}
  \end{subfigure}

  \caption{Image--question composition of the two benchmark data
  sources. The On-site Visual Data distribution measures the number
  of QA samples generated from each image, whereas the Accident
  Investigation Reports distribution measures the number of images
  required by each QA sample.}
  \label{fig:image-qa-composition}
\end{figure*}

\section*{Introduction}

Occupational injuries and work-related diseases impose substantial human and economic costs: the International Labour Organization estimates approximately 2.78 million work-related deaths annually and losses equivalent to 3.94\% of global GDP~\cite{ilo2017globalcost}. Reducing this burden requires reasoning across the safety lifecycle, from on-site inspection and compliance assessment to accident investigation, causal analysis, and preventive-control selection.

Recent multimodal large language models (MLLMs) have advanced general-purpose visual understanding and offer promising tools for industrial safety inspection and accident analysis~\cite{liu2023visual,wang2024qwen2vl,chen2024internvl,adil2025vlmhazard,ahmadi2025automatic}. Yet success on general-domain benchmarks does not show that a model can interpret a complete workplace scene, connect observed conditions to safety rules and accident knowledge, or recommend safe interventions.

Industrial hazards often cannot be judged from an isolated object or cropped view. A worker may wear the required PPE but still stand behind a reversing forklift~\cite{oshaForklift}, while a technician beside a stopped machine may be injured if it restarts during maintenance because its energy source was not isolated~\cite{osha1910147}. Recognizing such risks requires understanding the full scene, the task being performed, and how workers, equipment, and safeguards interact. Industrial safety is therefore a global, safety-critical reasoning problem rather than a collection of local recognition tasks.

Existing industrial-safety datasets predominantly evaluate local or isolated cues, such as PPE presence, unsafe actions, or individual violations~\cite{adil2025vlmhazard,chen2025constructionsite,abdullah2025isafetybench,li2025industryeqa,liu2026inspecsafe}. These tasks provide valuable perceptual foundations, but real-world safety warning requires combining evidence across the full scene: who is exposed, how equipment may move, which safeguards apply, and how the situation could develop into an accident. Current benchmarks provide limited support for this evidence-grounded progression from scene understanding to exposure assessment, accident-mechanism reasoning, and preventive action.

To address this gap, we introduce \textbf{SafeSceneReason}, a focused multimodal industrial-safety reasoning benchmark and companion training corpus built from two complementary sources: annotated workplace images from SH17 and DsLMF+~\cite{ahmad2025sh17,yang2023dslmf}, and more than 80,000 publicly available accident reports collected from OSHA, the NIOSH FACE program, and the U.S. Chemical Safety Board~\cite{osha2026investigations,niosh2024face,csb2026mission}. We categorize and screen the full report collection before retaining reports with sufficiently informative figures and contextual evidence for multimodal data construction. SafeSceneReason does not attempt to cover the entire occupational-safety lifecycle; instead, it targets a representative subset of tasks that connect workplace observations with compliance judgments, accident reasoning, and preventive decisions.

SafeSceneReason integrates two complementary components that correspond to two essential stages of industrial safety practice. The \emph{scene-centric component} supports proactive inspection before an incident occurs. Detection annotations are normalized into executable safety scene graphs containing objects, spatial relations, person--PPE links that associate each worker with the protective equipment worn by that individual, and rule-derived safety states. Declarative safety rules and executable programs generate reference answers for perception, counting, spatial reasoning, compliance assessment, compositional reasoning, causal reasoning, and counterfactual reasoning. A VLM may supplement restricted open-ended semantics such as attributes, actions, and relations, while an LLM may paraphrase questions; neither model is allowed to modify the program-derived answer.

The \emph{report-centric component} supports retrospective accident analysis and the transfer of lessons from previous incidents to future prevention. We use MinerU~\cite{wang2024mineru} to extract figures, captions, document context, and provenance metadata from NIOSH FACE and CSB reports~\cite{niosh2024face,csb2026mission}, and then construct questions through an evidence-constrained pipeline involving figure admission, evidence extraction, relational induction, reasoning-path composition, question--answer generation, and iterative quality assessment. Each evidence item is labeled according to whether it is observable from the image alone, requires textual clarification, or depends on the report text, preventing report-derived facts from being incorrectly presented as direct visual observations. Figure~\ref{fig:multi-image-machinery-example} illustrates a representative report-centric example in which evidence from four machinery-related figures is integrated through an explicit reasoning process to produce a concise safety conclusion.

The two components are necessary because either capability alone is incomplete. A model trained only on scene-level recognition may detect missing PPE or hazardous proximity without understanding the likely accident mechanism or selecting an effective control. Conversely, a model that can summarize historical reports may explain accidents after they occur but fail to ground that knowledge in a current workplace scene. By combining proactive scene inspection with retrospective accident knowledge, SafeSceneReason evaluates a more complete safety loop: observing conditions, judging compliance and exposure, anticipating incident development, and selecting preventive actions. All samples retain evidence references, intermediate reasoning representations, generation provenance, and verification results to support both model training and traceable evaluation.

Our main contributions are summarized as follows:
\begin{itemize}
    \item We introduce SafeSceneReason, a multimodal industrial-safety benchmark and training corpus grounded in real workplace scenes and official accident investigation materials. It evaluates a progression from perception and compliance assessment to hazard explanation, causal and counterfactual reasoning, and preventive decision-making.
    \item We develop two complementary and auditable data-synthesis pipelines corresponding to proactive inspection and retrospective accident learning. The scene-centric pipeline derives replayable answers from executable safety scene graphs, while the report-centric pipeline constructs evidence-grounded multi-step questions from accident figures and contextual report evidence.
    \item We establish strict quality-control and provenance mechanisms that separate factual inference from language generation, and systematically evaluate representative proprietary and open-source VLMs. The results reveal persistent gaps between visual recognition and reliable industrial-safety reasoning, particularly for compositional, causal, and intervention-oriented tasks.
\end{itemize}

\section*{Related Work}

\subsection{Visual and Multimodal Reasoning for Industrial Safety}

Early industrial-safety vision research mainly framed workplace inspection as recognition of directly observable objects, PPE, and unsafe actions, including worker localization, hard-hat and vest detection, fall detection, and abnormal-condition classification~\cite{ding2022safety,ahmad2025sh17,yang2023dslmf}. More recent resources extend this scope toward scene-level and multimodal understanding. VQA for Safety Compliance Checking introduces rule-conditioned question answering~\cite{ding2022safety}; ConstructionSite 10k combines scene descriptions, violation questions, rationales, and visual grounding~\cite{chen2025constructionsite}; and iSafetyBench, SteelBench, and InspecSafe-V1 evaluate safety understanding in industrial videos or multimodal inspection observations~\cite{abdullah2025isafetybench,yarrabothula2026steelbench,liu2026inspecsafe}. Vision--language models have also been studied for construction-hazard identification, while IndustryEQA evaluates embodied spatial, temporal, equipment, and worker-safety reasoning~\cite{adil2025vlmhazard,li2025industryeqa}.

These studies provide important perceptual and reasoning foundations, but most emphasize individual capabilities such as recognition, compliance classification, rationale generation, or grounding. They offer limited supervision for connecting full-scene evidence with worker exposure, applicable rules, plausible accident mechanisms, and preventive controls. SafeSceneReason complements them by evaluating representative stages of this progression and by linking workplace-scene observations with evidence from real accident investigations.

\subsection{Accident Investigation Report Analysis}

Accident reports have supported entity and cause--effect extraction, similar-incident retrieval, root-cause classification, preventive recommendation generation, retrieval-augmented analysis, and causal knowledge-graph construction~\cite{inoue2023safer,ahmadi2025automatic,patil2024improving,ren2025retrieval,chen2024information,liu2025large}. As summarized in Table~\ref{tab:unified_dataset_comparison}, these works generally treat reports as textual resources and address one task at a time. In contrast, our report-centric component jointly preserves documentary figures, captions, contextual narratives, and provenance from FACE and CSB investigations, then converts them into structured evidence graphs and verifiable multi-step reasoning paths.

\section*{Dataset Construction}

\begin{figure*}[t]
    \centering
    \includegraphics[width=\textwidth]{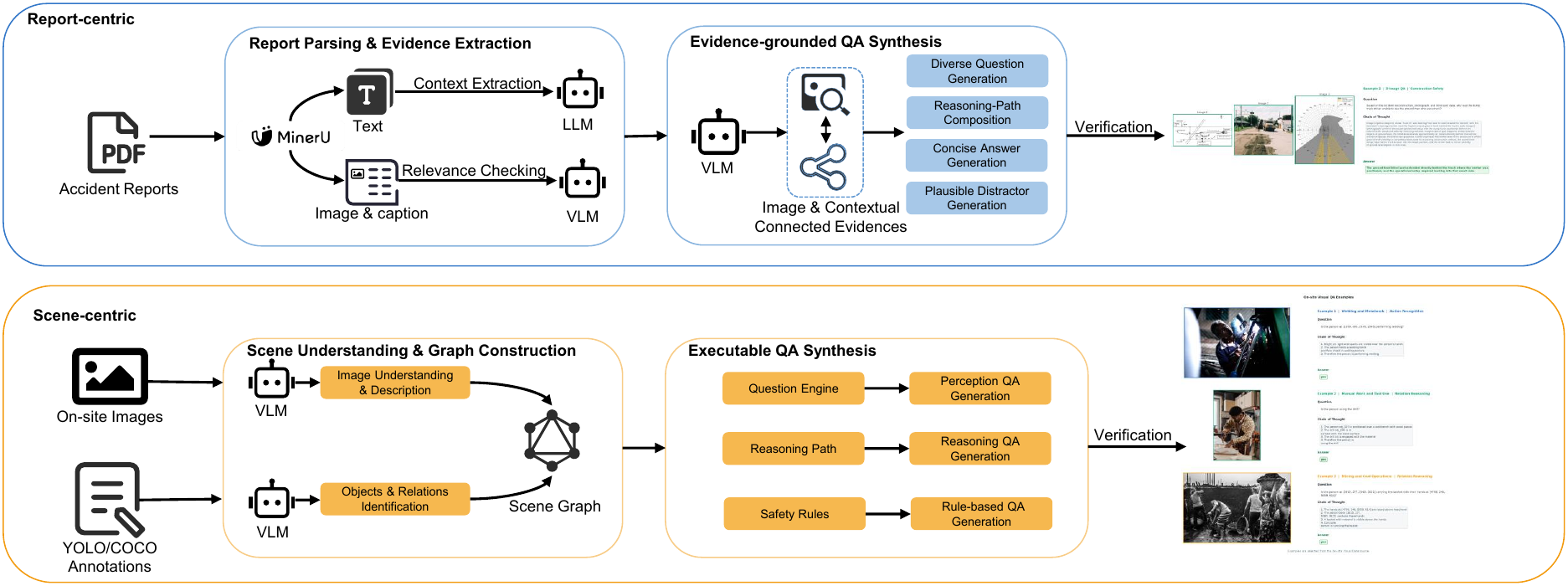}
    \caption{
    Overview of the SafeSceneReason framework.
    The framework consists of a scene-centric pipeline and a report-centric pipeline for constructing verified multimodal safety reasoning data.
    }
    \label{fig:safescenereason_pipeline}
\end{figure*}

Figure~\ref{fig:safescenereason_pipeline} summarizes the two complementary construction pipelines, while Figure~\ref{fig:dual_source_taxonomy} shows their taxonomy coverage. Both pipelines follow the same design principle: source-grounded evidence determines the answer, whereas generative models are used only where open-ended semantic interpretation or linguistic variation is required. Intermediate artifacts, source identifiers, and verification outcomes are retained so that each final sample can be traced back to its visual or documentary evidence.

\subsection{On-site Visual Data}

The scene-centric component uses industrial and construction datasets with annotations for workers, PPE, machinery, tools, activities, and hazardous conditions. We normalize YOLO- and COCO-format annotations~\cite{redmon2018yolov3,lin2014coco} into safety scene graphs whose nodes represent entities and whose edges encode geometric or functional relations such as \textit{wearing}, \textit{operating}, \textit{holding}, and \textit{inside a hazard zone}. Person--PPE associations are established using bounding-box containment and normalized spatial distance, allowing safety rules to be evaluated at the individual-worker level rather than at the image level.

Detection annotations provide the primary visual facts. A VLM may supplement restricted attributes, actions, and relations that are absent from the original labels, but every proposed item must refer to an existing graph entity, use a predefined predicate, and pass confidence and role-consistency checks. Declarative rules then derive compliance states and hazardous events. We adopt a conservative three-valued treatment of PPE: an item is marked \textit{missing} only when the relevant body region is visible and the required equipment is confidently absent; otherwise, insufficient evidence is retained as \textit{unknown}. This avoids converting detector failure or occlusion into a false violation.

Each question is paired with a program $z$ whose execution on scene graph $G$ yields
\begin{equation}
y = h\!\left(\operatorname{Exec}(z,G)\right),
\end{equation}
covering perception, counting, spatial, compliance, compositional, causal, and counterfactual reasoning. An LLM may paraphrase the question but cannot access or alter the program-derived answer. Verification replays every program, validates evidence references, checks question--predicate consistency, and removes duplicated, ungrounded, inconsistent, or visually unnecessary samples. We also bound the number of questions per category and balance positive and negative instances when the candidate pool permits. This pipeline produces 102,571 verified QA pairs from 8,099 SH17 images~\cite{ahmad2025sh17} and 8,010 pairs from 2,000 DsLMF+ images~\cite{yang2023dslmf}, for a total of 110,581 QA pairs over 10,099 images. Figure~\ref{fig:onsite-qa-per-image} summarizes the per-image QA distribution.

\subsection{Accident Investigation Reports}

The report-centric component begins with more than 80,000 publicly available accident reports collected from OSHA, NIOSH FACE, and CSB~\cite{osha2026investigations,niosh2024face,csb2026mission}. We categorize and screen the full collection, then retain reports containing sufficiently informative figures, captions, and contextual evidence for multimodal construction. MinerU~\cite{wang2024mineru} parses the retained reports into text, figures, captions, and page-level provenance. Figure admission removes logos, decorative elements, duplicated or incomplete crops, and figures lacking adequate context.

For each admitted sample, the pipeline proceeds through evidence extraction, relational induction, reasoning-path composition, question--answer generation, and quality assessment. Evidence items are explicitly labeled as directly observable from the figure, visually supported but requiring textual clarification, or available only from the report context. This information boundary prevents narrative facts from being presented as visual observations. The validated evidence graph is transformed into relational reasoning elements supporting operations such as comparison, quantitative interpretation, cross-evidence synthesis, causal analysis, and mitigation-oriented decision making. These elements are composed into explicit multi-step paths that specify the required premises and the permitted answer boundary before a question is written.

The generator preferentially produces four-option multiple-choice questions with one uniquely correct answer; it falls back to short-answer format when reliable distractors cannot be constructed. Candidates are checked for grounding, source consistency, reasoning coherence, answerability, information-boundary compliance, distractor validity, and difficulty. Invalid items are rejected, while repairable items are revised from structured verifier feedback and evaluated again. The resulting component contains 13,114 refined QA pairs with retained evidence, reasoning paths, provenance, and verification records. Figure~\ref{fig:report-images-per-qa} shows the number of documentary figures associated with each question, including both single-figure and multi-figure reasoning cases.

\section*{Experimental Analysis}

% ==================== QA-type performance table ====================
\begin{table*}[t]
\centering
\caption{
Accuracy (\%) of evaluated configurations across question types on the SafeSceneReason test set.
}
\label{tab:qa_type_performance}

\scriptsize
\setlength{\tabcolsep}{2.0pt}
\renewcommand{\arraystretch}{1.16}

\begin{tabular*}{\textwidth}{
    @{\extracolsep{\fill}}
    >{\raggedright\arraybackslash}p{3.05cm}
    c
    cccccccccc
    @{}
}
\toprule

\multirow{2}{*}{\textbf{Question Type}}
&
\multirow{2}{*}{\textbf{\#Q}}
&
\multirow{2}{*}{\makecell{\textbf{GPT-5.6}\\\textbf{sol}}}
&
\multirow{2}{*}{\makecell{\textbf{GPT-5.5}}}
&
\multirow{2}{*}{\makecell{\textbf{Gemini}\\\textbf{3.1 Pro}}}
&
\multirow{2}{*}{\makecell{\textbf{Gemini}\\\textbf{3.5 Flash}}}
&
\multirow{2}{*}{\makecell{\textbf{Claude}\\\textbf{Fable 5}}}
&
\multirow{2}{*}{\makecell{\textbf{Kimi}\\\textbf{K2.6}}}
&
\multirow{2}{*}{\makecell{\textbf{Qwen3.6}\\\textbf{27B}}}
&
\multicolumn{3}{c}{\textbf{Qwen3.5-9B}}
\\

\cmidrule(lr){10-12}

&
&
&
&
&
&
&
&
&
\textbf{Base}
&
\makecell{\textbf{SFT}\\\textbf{w/o CoT}}
&
\makecell{\textbf{CoT}\\\textbf{SFT}}
\\

\midrule

Quantitative Reasoning
& 279
& \second{87.81}
& 85.30
& 82.08
& 78.85
& 72.04
& 73.84
& 31.54
& 24.73
& 70.61
& \best{88.17}
\\

Spatial/Structural Reasoning
& 164
& \second{86.59}
& 84.15
& 75.00
& 76.22
& 73.17
& 65.24
& 51.22
& 45.73
& 75.61
& \best{87.20}
\\

Causal Reasoning
& 163
& \best{93.25}
& \second{90.80}
& 83.44
& 82.21
& 75.46
& 74.23
& 49.69
& 49.08
& 80.37
& \best{93.25}
\\

Comparative Reasoning
& 108
& 80.56
& \second{86.11}
& 76.85
& 78.70
& 67.59
& 71.30
& 26.85
& 25.00
& 71.30
& \best{92.59}
\\

Evidence Synthesis
& 101
& \second{91.09}
& \best{93.07}
& 80.20
& 76.24
& 73.27
& 73.27
& 42.57
& 36.63
& 74.26
& 85.15
\\

Hazard Identification
& 84
& \best{89.29}
& \second{85.71}
& 73.81
& 73.81
& 77.38
& 53.57
& 42.86
& 48.81
& 77.38
& 82.14
\\

Compliance Assessment
& 74
& 93.24
& \second{94.59}
& 86.49
& 82.43
& 82.43
& 78.38
& 47.30
& 35.14
& 82.43
& \best{95.95}
\\

Mitigation/Decision Making
& 27
& \second{92.59}
& \best{96.30}
& \second{92.59}
& 85.19
& 88.89
& 70.37
& 48.15
& 70.37
& 77.78
& 85.19
\\

\midrule

\textbf{Overall}
& \textbf{1,000}
& \second{88.70}
& 87.90
& 80.30
& 78.70
& 74.10
& 70.70
& 40.90
& 37.40
& 75.10
& \best{89.00}
\\

\textbf{Macro Average}
& --
& \second{89.30}
& \best{89.50}
& 81.31
& 79.21
& 76.28
& 70.02
& 42.52
& 41.94
& 76.22
& 88.70
\\

\bottomrule
\end{tabular*}

% \vspace{2pt}

\begin{minipage}{0.99\textwidth}
\scriptsize
\textbf{Note.} Overall and Macro Average denote micro- and category-macro accuracy. Best and second-best results are bold and underlined; SFT w/o CoT uses direct-answer supervision.
\end{minipage}

\end{table*}

\begin{figure}[t]
  \centering
  \includegraphics[width=\linewidth]{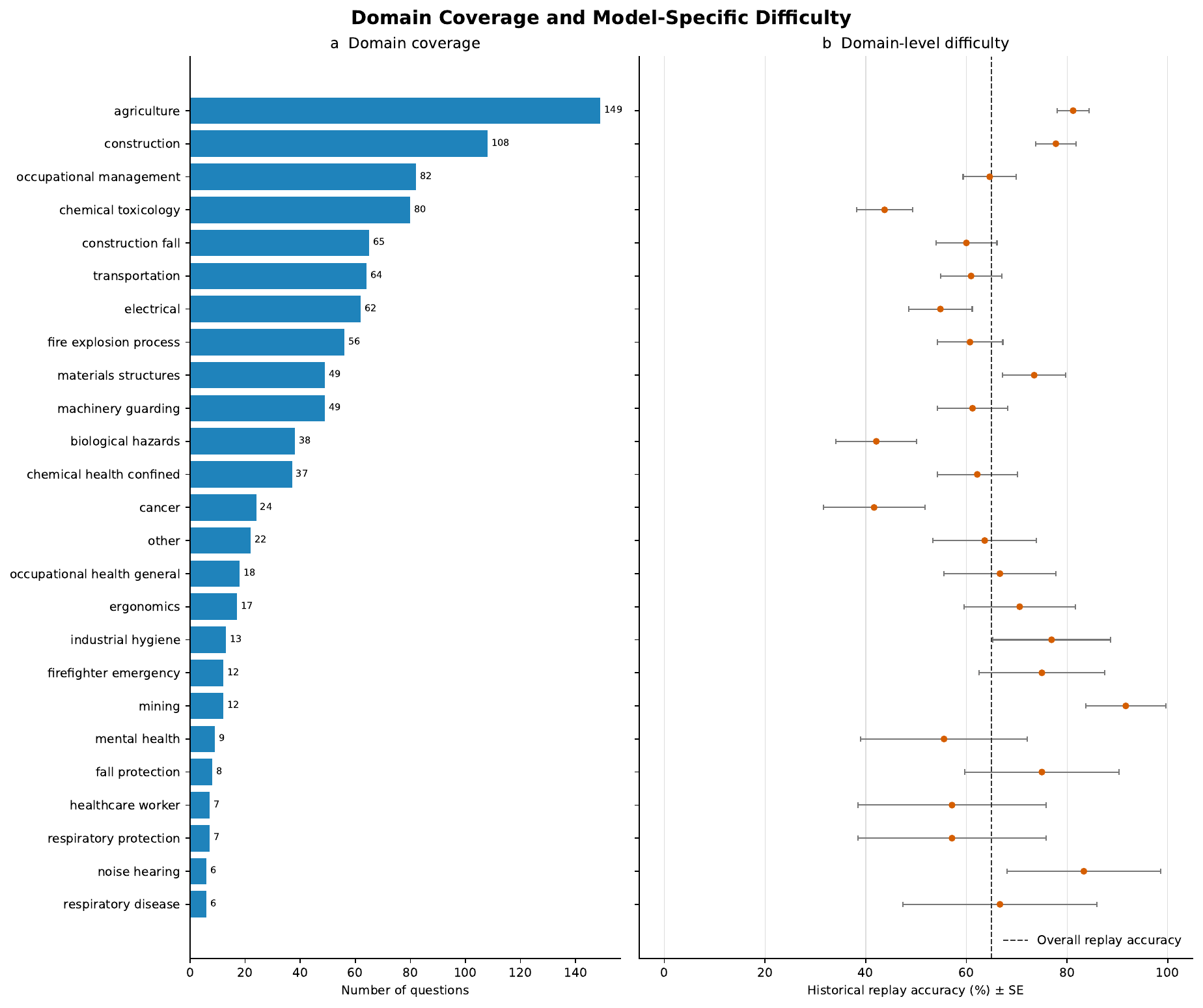}
  \caption{Domain coverage and Kimi-K2.6 difficulty under a historical replay. Error bars denote standard errors, and the dashed line marks the 65.0\% overall replay accuracy. This replay is used for domain-level diagnosis; Table~\ref{tab:qa_type_performance} reports the standardized benchmark results.}
  \label{fig:cdc-domain-coverage-difficulty}
\end{figure}

\subsection{Evaluation Setup}
\label{sec:experimental_setup}

We evaluate eight representative vision--language models---GPT-5.6-Sol, GPT-5.5, Gemini-3.1-Pro, Gemini-3.5-Flash, Claude-Fable-5, Kimi-K2.6, Qwen3.6-27B, and Qwen3.5-9B~\cite{openai2026gpt56,openai2026gpt55,google2026gemini31,google2026gemini35flash,anthropic2026fable5,moonshot2026kimi26,qwen2026qwen36,qwen2026qwen35}---together with two safety-domain adaptations of Qwen3.5-9B. The 1,000-example test set is balanced across answer positions and spans the eight reasoning categories in Table~\ref{tab:qa_type_performance}. All systems receive the same prompt and deterministic decoding configuration (temperature 0, top-$p=1$); a response is correct only when the parser extracts one option matching the reference, and ambiguous or unparseable outputs are counted as incorrect. We report micro accuracy over all examples and macro accuracy averaged equally across categories.

All local inference, fine-tuning, and replay analyses were conducted on a single server with eight NVIDIA A100 80GB GPUs. Proprietary models were accessed through their APIs under the same prompting and parsing protocol, whereas the Qwen models and fine-tuned checkpoints were evaluated locally using the same inference stack.

\subsection{Safety-Domain Fine-Tuning}
\label{sec:sft_setup}

We adapt Qwen3.5-9B using LoRA SFT~\cite{hu2022lora}. The answer-only split contains 8,810 training and 180 validation examples, while the CoT split contains 8,783 training and 185 validation examples; records matching the test set by sample ID, source-document ID, normalized question, or image path are removed, and the split is grouped by source document. The answer-only model emits a single option letter, whereas the CoT model emits a reasoning trace~\cite{wei2022chain} followed by the same letter. We freeze the visual encoder and multimodal aligner, apply rank-16 LoRA adapters ($\alpha=32$, dropout 0.05) to the language model, and train for two epochs in bfloat16 with sequence length 4,096, effective batch size 64, and AdamW~\cite{loshchilov2019decoupled} at $10^{-4}$. The checkpoint with the lowest validation loss is used for evaluation.

\subsection{Main Results}
\label{sec:overall_results}

\paragraph{Overall model comparison.}
Table~\ref{tab:qa_type_performance} reveals a pronounced capability hierarchy. CoT-SFT Qwen3.5-9B obtains the highest micro accuracy across all configurations (89.0\%), slightly exceeding GPT-5.6-Sol (88.7\%) and GPT-5.5 (87.9\%). GPT-5.5 achieves the best macro accuracy (89.5\%), followed by GPT-5.6-Sol (89.3\%) and CoT-SFT (88.7\%), indicating that the fine-tuned model is highly competitive overall but remains slightly less uniform across categories. Gemini-3.1-Pro and Gemini-3.5-Flash form a second tier at 80.3\% and 78.7\%, while Claude-Fable-5 and Kimi-K2.6 obtain 74.1\% and 70.7\%. The two untuned open Qwen models reach only 40.9\% and 37.4\%, producing a 51.3-point spread among base models and showing that general multimodal capability alone does not ensure reliable safety reasoning.

\paragraph{Category-level strengths are complementary.}
Among base models, GPT-5.6-Sol leads quantitative (87.8\%), spatial/structural (86.6\%), causal (93.3\%), and hazard-identification (89.3\%) tasks. GPT-5.5 instead leads comparative reasoning (86.1\%), evidence synthesis (93.1\%), compliance assessment (94.6\%), and mitigation/decision making (96.3\%). Thus, no proprietary model dominates every reasoning type: GPT-5.6-Sol is stronger on perception-linked and causal integration, whereas GPT-5.5 is stronger on cross-evidence comparison and decision-oriented tasks. Mid-tier models show larger category swings. Kimi-K2.6, for example, reaches 78.4\% on compliance assessment and 74.2\% on causal reasoning but only 53.6\% on hazard identification, suggesting that its errors depend strongly on task type and evidence form rather than on a single global capability level.

\paragraph{Safety-domain supervision closes most of the open-model gap.}
Answer-only SFT raises Qwen3.5-9B from 37.4\% to 75.1\% micro accuracy and from 41.9\% to 76.2\% macro accuracy. CoT SFT further reaches 89.0\% micro and 88.7\% macro accuracy, corresponding to gains of 51.6 and 46.8 points over the base model. The largest category improvements occur in comparative reasoning ($+67.6$ points), quantitative reasoning ($+63.4$), and compliance assessment ($+60.8$), where explicit intermediate supervision makes the required operations more visible. CoT-SFT also attains the best score on comparative reasoning (92.6\%), spatial/structural reasoning (87.2\%), and compliance assessment (96.0\%), and ties GPT-5.6-Sol on causal reasoning (93.3\%). However, it remains below the strongest proprietary models on evidence synthesis, hazard identification, and mitigation, showing that process supervision substantially improves composition without fully replacing broad domain knowledge and robust visual grounding.

\paragraph{Domain-specific difficulty under Kimi-K2.6.}
Figure~\ref{fig:cdc-domain-coverage-difficulty} provides a complementary diagnostic view using a historical Kimi-K2.6 replay with 65.0\% overall accuracy. Performance is relatively strong in well-represented agriculture (81.2\%, $n=149$) and construction (77.8\%, $n=108$), and remains high for materials and structures (73.5\%, $n=49$). In contrast, chemical toxicology (43.8\%, $n=80$), biological hazards (42.1\%, $n=38$), and cancer (41.7\%, $n=24$) are substantially below the overall replay rate; electrical safety is also challenging at 54.8\% ($n=62$). These gaps cannot be explained by sample frequency alone: chemical toxicology is one of the largest domains but remains among the most difficult, indicating a need for specialized terminology, technical evidence interpretation, and domain-specific causal knowledge. Very high scores in mining (91.7\%, $n=12$) and noise/hearing (83.3\%, $n=6$) have wide standard-error intervals and should therefore be interpreted cautiously rather than as evidence that these domains are intrinsically easy.

\subsection{Failure Mode Analysis}
\label{sec:failure_analysis}

\paragraph{Finding 1: Local recognition does not guarantee multi-evidence composition.}
The weakest open models collapse on quantitative, comparative, and evidence-synthesis questions: Qwen3.5-9B obtains 24.7\%, 25.0\%, and 36.6\%, while Qwen3.6-27B reaches 31.5\%, 26.9\%, and 42.6\%. Errors arise when several local observations must be aligned across figures or scene elements and transformed into one safety conclusion. This explains why models can identify individual objects yet still fail on the mechanism or control implied by their interaction.

\paragraph{Finding 2: Domain difficulty is not determined by representation frequency.}
Domains with similar support can differ sharply: materials/structures and machinery/guarding each contain 49 questions, yet Kimi-K2.6 reaches 73.5\% and 61.2\%, respectively; occupational management and chemical toxicology contain 82 and 80 questions but obtain 64.6\% and 43.8\%. The remaining gap therefore reflects domain-specific concepts and evidence demands, not merely class imbalance. Wide intervals in small domains further show that model confidence and benchmark uncertainty must be reported together.

\paragraph{Finding 3: Supervision reduces reasoning errors but leaves calibration failures.}
Although answer positions are balanced, base Qwen3.5-9B achieves 52.4\% when option A is correct but only 31.6--33.6\% for the other positions, revealing a response-format shortcut. CoT SFT greatly improves procedural reasoning, yet its lowest categories remain hazard identification (82.1\%), evidence synthesis (85.1\%), and mitigation/decision making (85.2\%). Explicit traces therefore reduce composition errors but do not fully solve decisive-evidence grounding, uncertainty calibration, or safe intervention selection.

\section*{Conclusion}

We introduced \textbf{SafeSceneReason}, a multimodal benchmark and companion training corpus for connecting observable workplace conditions with knowledge from real accident investigations. Its scene-centric pipeline converts annotated workplace images into executable safety scene graphs and derives replayable answers through program execution, while its report-centric pipeline constructs evidence-grounded questions from documentary figures, contextual text, and explicit multi-step reasoning paths. Across representative proprietary and open-source vision--language models, the results reveal a clear separation between general visual recognition and dependable safety reasoning. Strong models perform well on several perception and compliance tasks, yet comparative, quantitative, and multi-evidence questions remain challenging. Safety-domain supervision substantially improves Qwen3.5-9B, with chain-of-thought training providing larger and more consistent gains than direct-answer supervision.

SafeSceneReason is intended as a focused step rather than a complete representation of the occupational-safety lifecycle. Its coverage is bounded by the available workplace annotations, selected accident-report sources, and the safety concepts that can be supported by traceable visual and textual evidence. The observed domain-level variation and persistent failures in evidence integration, calibration, and intervention selection indicate that high aggregate accuracy alone is insufficient for safety-critical use. Future work should broaden coverage to temporal workplace activities, video and sensor evidence, additional industries and accident repositories, and expert-centered evaluation of explanations and recommended controls. We hope the benchmark supports the development of multimodal systems whose safety judgments are not only accurate, but also grounded, auditable, and appropriately cautious.

\bigskip

\bibliography{aaai2027}


\end{document}